\documentclass{article}
\usepackage{ijcai18}

\usepackage{times}
\usepackage{xcolor}
\usepackage{soul}
\usepackage[utf8]{inputenc}
\usepackage[small]{caption}

\usepackage{subcaption}
\usepackage{helvet}
\usepackage{courier}
\usepackage{multirow}
\usepackage{xcolor}
\usepackage{amsmath,amscd,amsbsy,amssymb,latexsym,url,bm}
\usepackage{epsfig,epstopdf,graphicx}
\usepackage{enumitem,balance}
\usepackage{wrapfig}
\usepackage{mathrsfs, euscript}
\usepackage{algorithm}
\usepackage[colorinlistoftodos]{todonotes}
\usepackage{algpseudocode}
\usepackage{ifpdf}
\usepackage{makecell}
\usepackage{tikz}
\usetikzlibrary{matrix,chains,positioning,decorations.pathreplacing,arrows}
\usepackage{amsthm}

\newtheorem{theorem}{Theorem}
\newtheorem{definition}{Definition}
\newcommand{\argmax}[1]{\underset{#1}{\operatorname{arg}\operatorname{max}}\;}
\newcommand{\argmin}[1]{\underset{#1}{\operatorname{arg}\operatorname{min}}\;}

\title{Learning to Design Games: Strategic Environments \\in Reinforcement Learning\thanks{Accepted by the 27th International Joint Conference on Artificial Intelligence (IJCAI-18).}}

\author{
Haifeng Zhang$^1$\thanks{This work is done during Haifeng Zhang's visit at UCL. Jun Wang and Weinan Zhang are the corresponding authors of this paper.}, 
Jun Wang$^2$, 
Zhiming Zhou$^3$, 
Weinan Zhang$^3$,
Ying Wen$^2$,
Yong Yu$^3$,
Wenxin Li$^1$
\\ 
$^1$ Peking University\\
$^2$ University College London\\
$^3$ Shanghai Jiao Tong University  \\
pkuzhf@pku.edu.cn,
jun.wang@cs.ucl.ac.uk,
wnzhang@sjtu.edu.cn
}

\begin{document}

\maketitle
\begin{abstract}
In typical reinforcement learning (RL), the environment is assumed given and the goal of the learning is to identify an optimal policy for the agent taking actions through its interactions with the environment.  In this paper, we extend this setting by considering the environment is not given, but controllable and learnable through its interaction with the agent at the same time. This extension is motivated by environment design scenarios in the real-world, including game design, shopping space design and traffic signal design. Theoretically, we find a dual Markov decision process (MDP) w.r.t. the environment to that w.r.t. the agent, and derive a policy gradient solution to optimizing the parametrized environment. Furthermore, discontinuous environments are addressed by a proposed general generative framework. Our experiments on a Maze game design task show the effectiveness of the proposed algorithms in generating diverse and challenging Mazes against various agent settings.
\end{abstract}

\section{Introduction} \label{sec:intro}

Reinforcement learning (RL) is typically concerned with a scenario where an agent (or multiple agents) taking actions and receiving rewards from an environment \cite{kaelbling1996reinforcement}, and the goal of the \emph{learning} is to find an optimal policy for the agent that maximizes the cumulative reward when interacting with the environment.
Successful applications include playing games \cite{mnih2013playing,silver2016mastering}, scheduling traffic signal  \cite{abdulhai2003reinforcement}, regulating ad bidding \cite{cai2017real}, to name just a few. 

In most RL approaches, such as SARSA and Q-learning \cite{sutton1998reinforcement}, the model of the environment is, however, not necessarily known a priori before learning the optimal policy for the agent. Alternatively,  model-based approaches, such as DYNA \cite{sutton1990integrated} and prioritized sweeping \cite{moore1993prioritized}, require establishing the environment model while learning the optimal policy. Nonetheless, in either case, the environment is assumed given and mostly either stationary or non-stationary without a purposive control \cite{kaelbling1996reinforcement}.

In this paper, we extend the standard RL setting by considering the environment is strategic and controllable. We aim at learning to design an environment via interacting with an also learnable agent or multiple agents. This has many potential applications, ranging from designing a game (environment) with a desired level of difficulties in order to fit the current player's learning stage \cite{togelius2008experiment} and designing shopping space to impulse customers purchase and long stay \cite{penn2005complexity} to controlling traffic signals \cite{ceylan2004traffic}. In general, we propose and formulate the design problem of environments which interact with intelligent agents/humans. We consider designing these environments via machine learning would release human labors and benefit social efficiency. 
Comparing to the well-studied image design/generation problem \cite{goodfellow2014generative}, environment design problem is new in three aspects: 
(i) there is no ground-truth samples; 
(ii) the sample to be generated may be discontinuous; 
(iii) the evaluation of a sample is through learning intelligent agents. 

Our formulation extends the scope of RL by focusing on the environment modeling and control. Particularly, in an adversarial case, on one hand, the agent aims to maximize its accumulative reward; on the other hand, the environment tends to minimize the reward for a given optimal policy from the agent. This effectively creates a minimax game between the agent and the environment. Given the agent's playing environment MDP, we, theoretically, find a dual MDP w.r.t. the environment, i.e., how the environment could decide or sample the successor state given the agent's current state and an action taken.
Solving the dual MDP yields a policy gradient solution \cite{williams1992simple} to optimize the parametric environment achieving its objective. When the environment's parameters are not continuous, we propose a generative modeling framework for optimizing the parametric environment, which overcomes the constraints on the environment space. Our experiments on a Maze game generation task show the effectiveness of generating diverse and challenging Mazes against various types of agents in different settings. We show that our algorithms would be able to successfully find the weaknesses of the agents and play against them to generate purposeful environments. 

The main contributions of this paper are threefold:
(i) we propose the environment design problem, which is novel and potential for practical applications; 
(ii) we reduce the problem to the policy optimization problem for continuous cases and propose a generative framework for discontinuous cases; 
(iii) we apply our methods to Maze game design tasks and show their effectiveness by presenting the generated non-trivial Mazes. 


\section{Related Work}

Reinforcement learning (RL) \cite{sutton1998reinforcement} studies how an intelligent agent learns to take actions through the interaction with an environment over time. In a typical RL setting, the environment is unknown yet fixed, and the focus is on optimizing the agent policies. Deep reinforcement learning (DRL) is a marriage of deep neural networks \cite{lecun2015deep} and RL; it makes use of deep neural networks as a function approximator in the decision-making framework of RL to achieve human-level control and general intelligence \cite{mnih2015human}. 
In this paper, instead, we consider a family of problems that is an extension of RL by considering that the environment is controllable and strategic. Unlike typical RL, our subject is the strategic environment not the agent, and the aim is to learn to design an optimal (game) environment via the interaction with the intelligent agent. 

Our problem of environment design is related to the well-known mechanism design problem\cite{nisan2001algorithmic}, which studies how to design mechanisms for participants that achieves some objectives such as social welfare. In most studies, the designs are manual. Our work focuses on automated environment (mechanism) design by machine learning. Thus, we formulate the problem based on MDP and provide solutions based on RL. In parallel, the automated game-level design is a well-studied problem by applying search-based procedural content generation\cite{togelius2011search}. For generating game-levels that conform to design requirements, genetic algorithm (GA) is proposed as a searcher. Our work instead providing sound solutions based on RL methods, which bring new properties such as gradient direction searching and game feature learning. 

In the field of RL, our problem is related to safe/robust reinforcement learning, which maximizes the expectation of the return under some safety constraints such as uncertainty \cite{garcia2015comprehensive,morimoto2005robust}, due to the common use of parametric MDPs. However, our problem setting is entirely different from safe RL as their focus is on single agent learning in an unknown environment, whereas our work is concerned with the learning of the environment to achieve its own objective. Our problem is also different from agent reward design \cite{sorg2010reward}, which optimizes designer's cumulative reward given by a fixed environment (MDP). However, the environment is learnable in our setting.
Another related work, FeUdal networks \cite{vezhnevets2017feudal}, introduces transition policy gradient to update the proposed manager model, which is a component of agent policy. This is different from our transition gradient which is for updating the environment.

Our formulation is a general one, applicable in the setting where there are multiple agents \cite{busoniu2008comprehensive}. It is worth mentioning that although multi-agent reinforcement learning (MARL) studies the strategic interplays among different entities, the game (either collaborative or competitive) is strictly among multiple agents \cite{littman1994markov,hu2003nash}. By contrast, the strategic interplays in our formulation are between an agent (or multiple agents) and the environment. The recent work, interactive POMDPs \cite{gmytrasiewicz2005framework}, aims to spread beliefs over physical states of the environment and over models of other agents, but the environment in question is still non-strategic. Our problem, thus, cannot be formulated directly using MARL as the decision making of the environment is in an episode-level, while policies of agents typically operate and update in each time-step within an episode. 

In addition, our minimax game formulation can also be found in the recently emerged generative adversarial nets (GANs), where a generator and a discriminator play a minimax adversarial game \cite{goodfellow2014generative}. Compared to GANs, our work addresses a different problem, where the true samples of desired environments are missing in our scenario; the training of our environment generator is guided by the behaviours of the agent (corresponding the GAN discriminator) who aims to maximize its cumulative reward in a given environment.



\section{RL with Controllable Environment} \label{sec:theory}

\subsection{Problem Formulation}

Let us first consider the standard reinforcement learning framework. In this framework there are a learning agent and a Markov decision process (MDP) $\mathcal{M} = \langle \mathcal{S}, \mathcal{A}, \mathcal{P}, \mathcal{R}, \gamma\rangle$, where $\mathcal{S}$ denotes state space, $\mathcal{A}$ action space, $\mathcal{P}$ state transition probability function, $\mathcal{R}$ reward function and $\gamma$ discounted factor. The agent interacts with the MDP by taking action $a$ in state $s$ and observing reward $r$ in each time-step, resulting in a trajectory of states, actions and rewards: $H_{1\ldots \infty}= \langle S_1,A_1,R_1,S_2,A_2,R_2\ldots \rangle, S_t \in \mathcal{S}, A_t \in \mathcal{A}, R_t \in \mathbb{R}$, where $\mathbb{P}[S_{t+1} = s'|S_t = s, A_t = a] = \mathcal{P}(s, a, s')$ and $\mathbb{E}[R_t|S_t = s, A_t = a] = \mathcal{R}(s, a)$ hold.\footnote{In this paper, we use $S_t, A_t, R_t$ when they are in trajectories while using $s, a, r$ otherwise.} The agent selects actions according to a policy $\pi_\phi$, where $\pi_\phi(a|s)$ defines the probability that the agent selects action $a$ in state $s$. The agent learns $\pi_\phi$ to maximize the return (cumulative reward) $G=\sum_{t=1}^\infty \gamma^{t-1} R_t$. 


In the standard setting, the MDP is given fixed while the agent is flexible with its policy to achieve its objective. We extend this setting by also giving flexibility and purpose to $\mathcal{M}$. Specifically, we parametrize $\mathcal{P}$ as $\mathcal{P}_\theta$ and set the objective of the MDP as $O(H)$, which can be arbitrary based on the agent's trajectory. We intend to design (generate) an MDP that achieves the objective along with the agent achieving its own objective:
\begin{align}
\theta^* = \argmax{\theta} \mathbb{E}\big[O(H)|\mathcal{M}_\theta = \langle \mathcal{S}, \mathcal{A}, \mathcal{P}_\theta, \mathcal{R}, \gamma\rangle; \nonumber \\ 
\pi_{\phi^*} = \argmax{\pi_\phi} \mathbb{E}[G|\pi_\phi; \mathcal{M}_\theta] \big].
\label{eq:general-problem} 
\end{align} 

\subsubsection{Adversarial Environment}

In this paper, we consider a particular objective of the environment that it acts as an adversarial environment minimizing the expected return of the single agent, i.e., $O(H) = \sum_{t=1}^\infty - \gamma^{t-1}R_t = - G$.
This adversarial objective is useful in the game design domain because for many games the game designer need to design various game levels or set various game parameters to challenge game players playing with various game strategies. Thus, the relationship between the environment(game) and the agent(player) are adversarial. We intend to transfer this design work from human to machine by applying appropriate machine learning methods. Formally, the objective function is formulated as:
\begin{align}
\theta^* = \argmin{\theta} \max_\phi \mathbb{E}[G|\pi_\phi;  \mathcal{M}_\theta = \langle \mathcal{S}, \mathcal{A}, \mathcal{P}_\theta, \mathcal{R}, \gamma\rangle].
\label{eq:adversarial-problem}
\end{align}

\begin{figure}[t]
	\centering
	\includegraphics[width=1\columnwidth]{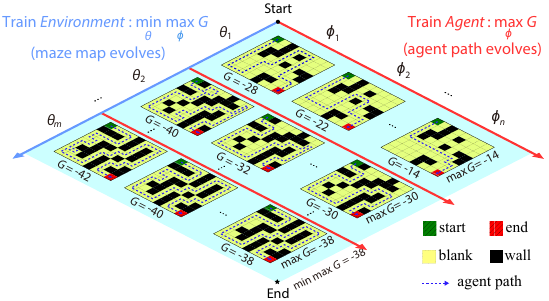}
	\caption{An example of adversarial Maze design. The detailed definition of the Maze environment is provided in Sec.\ref{sec:maze}. In short, an agent tries to find the shortest path from the start to the end in a given Maze map, while the Maze environment tries to design a map to make the path taken by the agent longer. In the direction of $\phi$, the parameter of an agent policy evolves, whereas in the direction of $\theta$, the parameter of the Maze environment evolves. The cumulative reward $G$ is defined as the opposite number of the length of the path.}\label{fig:env-gan-iteration}
	\vspace{-10pt}
\end{figure}

In general, we adopt an iterative framework for learning $\theta$ and $\phi$. In each iteration, the environment updates its parameter to maximize its objective w.r.t. the current agent policy then the agent updates its policy parameter by taking sufficient steps to be optimal w.r.t. the updated environment, as illustrated by Fig.~\ref{fig:env-gan-iteration} for learning the environment of a Maze. Since the agent's policy can be updated using well-studied RL methods, we focus on the update methods for the environment. In each iteration, given the agent's policy parameter $\phi^*$, the objective of the environment is

\begin{align} \label{eq:env-design-fix-agent}
\theta^* = \argmin{\theta} \mathbb{E}[G|\mathcal{M}_\theta = \langle \mathcal{S}, \mathcal{A}, \mathcal{P}_\theta, \mathcal{R}, \gamma\rangle; \pi_{\phi^*}].
\end{align}

In the following sections, we propose two methods to solve this problem for continuous and discontinuous environments. 




\subsection{Gradient Method for Continuous Environment} \label{sec:trans-grad}

In this section, we propose a gradient method for continuous environment, i.e. the value of the transition probability for any $\langle s, a, s' \rangle$ can be arbitrary in $[0, 1]$. Thus, the parameter $\theta$ of the environment actually consists of the values of the transition function $\mathcal{P}(s, a, s')$ for each $\langle s, a, s' \rangle$. Our task is to optimize the values of the transition function to minimize the agent's cumulative reward.

To update the environment, we try to find the gradient of the environment objective w.r.t. $\theta$. We derive the gradient by 
taking a new look at the environment and the agent in the opposite way, that the original environment $\mathcal{M}^A$ as an agent and the original agent as a part of the new environment $\mathcal{M}^E$. Viewing in this way, the original environment $\mathcal{M}^A$ takes action $A^E_t$ to determine the next state $S^A_{t+1}$ given the current state $S^A_t$ and the agent's action $A^A_t$. Thus we define the state $s^E$ in $\mathcal{M}^E$ as the combination $\langle s^A, a^A \rangle$. On the other hand, given the original environment's action $A^E_t =S^A_{t+1}$ , the agent policy $\pi_{\phi^*}^A(s^A)$ acts as a transition in $\mathcal{M}^E$ to determine $A^A_{t+1}$ as part of the next state $S^E_{t+1}=\langle S^A_{t+1}, A^A_{t+1} \rangle$ in $\mathcal{M}^E$. Furthermore, optimizing agent policy in $\mathcal{M}^E$ is equal to optimizing environment transition in $\mathcal{M}^A$. 

Theoretically, we reduce our transition optimization problem in Eq.~(\ref{eq:env-design-fix-agent}) to the well-studied policy optimization problem through a proposed concept of a duel MDP-policy pair.

\begin{definition}[Duel MDP-policy pair]

For any MDP-policy pair $\langle \mathcal{M}^A, \pi^A \rangle$, where $\mathcal{M}^A = \langle \mathcal{S}^A, \mathcal{A}^A, \mathcal{P}^A, \mathcal{R}^A, \gamma^A \rangle$ with start state distribution $p_1^A$ and terminal state set $\mathcal{S}_T^A$, there exists a dual MDP-policy pair $\langle \mathcal{M}^E, \pi^E \rangle$, where $\mathcal{M}^E = \langle \mathcal{S}^E, \mathcal{A}^E, \mathcal{P}^E, \mathcal{R}^E, \gamma^E \rangle$ with start state distribution $p_1^E$ and terminal action set $\mathcal{A}_T^E$ satisfying:
\begin{itemize}
\item $\mathcal{S}^E = \mathcal{S}^A \times \mathcal{A}^A = \{\langle s^A, a^A\rangle | s^A \in \mathcal{S}^A, a^A \in \mathcal{A}^A\}$, a state in $\mathcal{M}^E$ corresponds to a combination of successive state and action in $\mathcal{M}^A$;
\item $\mathcal{A}^E = \mathcal{S}^A = \{s^A | s^A \in \mathcal{S}^A\}$, an action in $\mathcal{M}^E$ corresponds to a state in $\mathcal{M}^A$;
\item $\mathcal{P}^E(s_i^E, a^E, s_{i'}^E) = \mathcal{P}^E(\langle s_j^A, a_k^A\rangle, s^A, \langle s_{j'}^A, a_{k'}^A\rangle) = \begin{cases}
		\pi^A(a_{k'}^A|{s^A}) & s^A=s_{j'}^A\\
		0 & s^A \neq s_{j'}^A
		\end{cases}$
		, the transition in $\mathcal{M}^E$ depends on the policy in $\mathcal{M}^A$;
\item $\mathcal{R}^E(s_i^E, a^E) = \mathcal{R}^E(\langle s_j^A, a_k^A\rangle, s^A) = \mathcal{R}^A(s_j^A, a_k^A)$, the rewards in $\mathcal{M}^E$ are the same as in $\mathcal{M}^A$;
\item $\gamma^E = \gamma^A$, the discounted factors are the same;
\item $p_1^E(s^E) = p_1^E(\langle s^A, a^A \rangle) = p_1^A(s^A) \pi^A(a^A|s^A)$, start state distribution in $\mathcal{M}^E$ depends on start state distribution and the first action distribution in $\mathcal{M}^A$;
\item $\mathcal{A}_T^E = \{s^A | s^A \in \mathcal{S}_T^A \}$, terminal action in $\mathcal{M}^E$ corresponds to terminal state in $\mathcal{M}^A$;
\item $\pi^E(a^E|s^E) = \pi^E(s_{i'}^A|\langle s_i^A, a^A \rangle) = \mathcal{P}^A(s_i^A, a^A, s_{i'}^A)$, policy in $\mathcal{M}^E$ corresponds to transition in $\mathcal{M}^A$.
\end{itemize}

\end{definition}

We can see that the dual MDP-policy pair in fact describes an equal mechanism as the original MDP-policy pair from another perspective. 
Based on the dual MDP-policy pair, we give three theorems to derive the gradient of the transition function. The proofs are omitted for space reason.

\begin{theorem}\label{thm:pdf}
For an MDP-policy pair $\langle \mathcal{M}^A, \pi^A \rangle$ and its duality $\langle \mathcal{M}^E, \pi^E \rangle$, the distribution of trajectory generated by $\langle \mathcal{M}^A, \pi^A \rangle$ is the same as the distribution of a bijective trajectory generated by $\langle \mathcal{M}^E, \pi^E \rangle$, i.e. $\mathbb{P}[H^A|M^A,\pi^A]=\mathbb{P}[H^E|M^E,\pi^E]$, where $H^E=b(H^A), H^A=b^{-1}(H^E)$.
\end{theorem}


\begin{theorem} \label{thm:exp-return}
For an MDP-policy pair $\langle \mathcal{M}^A, \pi^A \rangle$ and its duality $\langle \mathcal{M}^E, \pi^E \rangle$, the expected return of two bijective state-action trajectories, $H^A=b^{-1}(H^E)$ from $\langle \mathcal{M}^A, \pi^A \rangle$ and $H^E=b(H^A)$ from $\langle \mathcal{M}^E, \pi^E \rangle$, are equal.
\end{theorem}

\begin{theorem} \label{thm:exp-return-all}
For an MDP-policy pair $\langle \mathcal{M}^A, \pi^A \rangle$ and its duality $\langle \mathcal{M}^E, \pi^E \rangle$, the expected return of $\langle \mathcal{M}^A, \pi^A \rangle$ is equal to the expected return of $\langle \mathcal{M}^E, \pi^E \rangle$, i.e., $\mathbb{E}[G^A|\pi^A, \mathcal{M}^A] = \mathbb{E}[G^E|\pi^E, \mathcal{M}^E]$.
\end{theorem}

Theorem~\ref{thm:exp-return} can be understood by the equivalence between $H^A$ and $H^E$ and the same generating probability of them as given in Theorem~\ref{thm:pdf}. Theorem~\ref{thm:exp-return-all} naturally extends Theorem~\ref{thm:exp-return} from the single trajectory to the distribution of trajectory according to the equal probability mass function given by Theorem~\ref{thm:pdf}.

Now we consider $\langle \mathcal{M}^A_\theta, \pi^A \rangle$ and its duality $\langle \mathcal{M}^E, \pi^E_\theta \rangle$, where $\mathcal{P}^A_\theta$ and $\pi^E_\theta$ are of the same form about $\theta$. Given $\theta$, $\mathcal{P}^A_\theta$ and $\pi^E_\theta$ are exactly the same, resulting in $\mathbb{E}[G^A|\pi^A, \mathcal{M}^A_\theta] = \mathbb{E}[G^E|\pi^E_\theta, \mathcal{M}^E]$ according to Theorem~\ref{thm:exp-return-all}. Thus optimizing $\mathcal{P}^A_\theta$ as Eq.~(\ref{eq:env-design-fix-agent}) is equivalent to optimizing $\pi^E_\theta$:
\begin{align}
\theta^* = \argmin{\theta} \mathbb{E}[G|\mathcal{M}^A_\theta;\pi^A_{\phi^*}] = \argmin{\theta} \mathbb{E}[G|\pi^E_\theta; \mathcal{M}^E_{\phi^*}].
\end{align}
We then apply the policy gradient theorem \cite{sutton1999policy} on $\pi^E_\theta$ and derive the gradient for $\mathcal{P}^A_\theta$:
\begin{equation}\begin{aligned}
\nabla_\theta J(\theta) 
&= \mathbb{E}[\nabla_\theta \log \pi^E_\theta (a^E|s^E) Q^E(s^E,a^E) | \pi^E_\theta; \mathcal{M}^E_{\phi^*}] \\
&= \mathbb{E}[\nabla_\theta \log \mathcal{P}^A_\theta (s_i^A, a^A, s_{i'}^A) V^A(s_{i'}^A) | \mathcal{M}^A_\theta; \pi^A_{\phi^*}], \label{eq:transition-gradient}
\end{aligned}
\end{equation}
where $J(\theta)$ is cost function, $Q^E(s^E,a^E)$ and $V^A(s_{i'}^A)$ are action-value function and value function of $\langle \mathcal{M}^E, \pi^E_\theta \rangle$ and $\langle \mathcal{M}^A_\theta, \pi^A \rangle$ respectively; and can be proved equal due to the equivalence of the two MDPs.

We name the gradient in Eq.~(\ref{eq:transition-gradient}) as \emph{transition gradient}. Transition gradient can be used to update the transition function in an iterative way. In theory, it performs as well as policy gradient since it is equivalent to the policy gradient in the circumstance of the dual MDP-policy pair. 

\subsection{Generative Framework for Discontinuous Environment} \label{sec:rl-gen}

\begin{figure}[t]
	\centering
	\includegraphics[width=1\columnwidth]{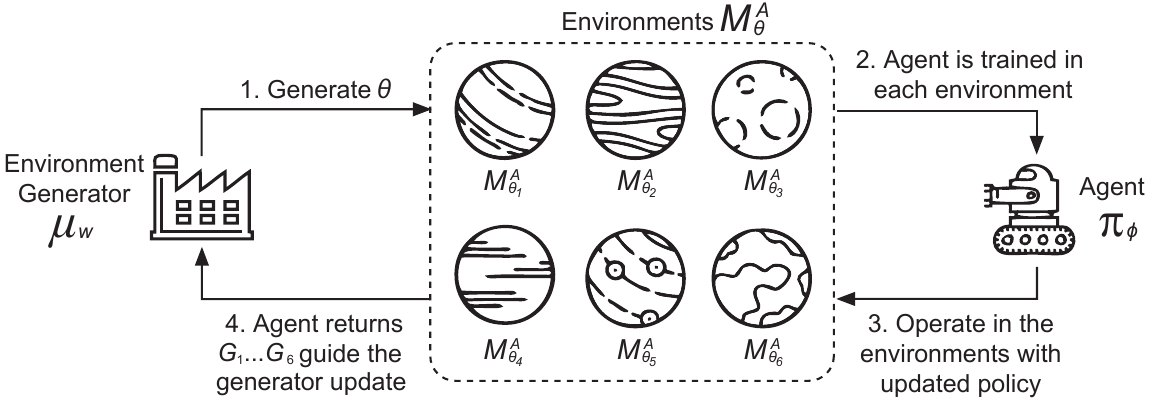}
	\caption{Framework dealing with discontinuous environment. Generator generates environment parameter $\theta$. For each $\mathcal{M}^A_{\theta}$, the agent policy is trained. Then the policy is tested in the generated environments and the returns are observed, which finally guide the generator to update.}\label{fig:env-gan-frame}
	\vspace{-10pt}
\end{figure}

The transition gradient method proposed in the last section only works for continuous environment. For discontinuous environment, i.e. the range of the transition function $\mathcal{P}(s, a, s')$ is not continuous in $[0, 1]$, we cannot directly take the gradient of the transition function w.r.t. $\theta$.

To deal with the discontinuous situation, we propose a generative framework to find the optimal $\theta$ alternative to the gradient method. In general, we build a parametrized generator to generate a distribution of the environment, then update the parameter of the generator by evaluating the environments it generates (illustrated in Fig.~\ref{fig:env-gan-frame}). Specifically, we generate environment parameter $\theta$ using a $w$-parametrized generator $\mu_w$, then optimize $w$ to obtain the (local) optimal $w^*$ and a corresponding optimal distribution of $\theta$. 
Formally, our optimization objective is formulated as
\begin{align}
w^* = \argmin{w} &\mathbb{E}_{\theta \sim \mu_w} \big[\mathbb{E}[G| \nonumber \\
&\mathcal{M}^A_\theta = \langle \mathcal{S}^A, \mathcal{A}^A, \mathcal{P}^A_\theta, \mathcal{R}^A, \gamma^A \rangle; \pi_{\phi^*}] \big]. \label{eq:generator-obj}
\end{align}
We model the generation process using an auxiliary MDP $\mathcal{M}^{\mu}$, i.e., the generator $\mu_w$ generates $\theta$ and updates $w$ in a reinforcement learning way. The reason we adopt reinforcement learning other than supervised learning is that in this generative task, (i) there is no training data to describe the distribution of the desired environments so we cannot compute likelihood of generated environments and (ii) we can only evaluate a generated environment through sampling, i.e., performing agents in the generated environment and getting a score from the trajectory, which can be naturally modeled by reinforcement learning by viewing the score as a reward of the actions of the generator.

In detail, the generator $\mu_w$ consists of three elements $\langle \mathcal{M}^{\mu}, \pi^{\mu}_w, f^{\mu} \rangle$. For generating $\theta$, an auxiliary agent with policy $\pi^{\mu}_w$ acts in $\mathcal{M}^{\mu}$ to generate a trajectory $H^{\mu}$, after that $\theta$ is determined by the transforming function $\theta = f^{\mu}(H^{\mu})$, i.e., the distribution of $\theta$ is based on the distribution of trajectories, which are further induced by playing $\pi^{\mu}_w$ in $\mathcal{M}^{\mu}$. For adversarial environments, the reward of the generator is designed to be opposite to the return of the agent got in $\mathcal{M}_\theta$, which reflects the minimization objective in Eq.~(\ref{eq:generator-obj}). Thus, $w$ can be updated by applying policy gradient methods on $\pi^{\mu}_w$. 

There are various ways to designing $\mathcal{M}^{\mu}$ for a particular problem. Here we provide a general design that can be applied to any environment. Briefly, we generate the environment parameter in an additive way and ensures the validity along the generation process. In detail, we reshape the elements of $\theta$ as a vector $\theta = \langle x_1, x_2, \ldots , x_{N_\theta} \rangle, x_k \in X_k$ and design $\mathcal{M}^{\mu} = \langle \mathcal{S}^{\mu}, \mathcal{A}^{\mu}, \mathcal{P}^{\mu}, \mathcal{R}^{\mu}, \gamma^{\mu}=1 \rangle$ to generate $\theta$:

\begin{itemize}
	\item $\mathcal{S}^{\mu} = \{v_k = \langle x_1, x_2, \ldots , x_k \rangle |k = 0\ldots N_\theta, \exists v_{N_\theta} = \langle x_1, x_2, \ldots , x_k, x_{k+1}' \ldots x_{N_\theta}' \rangle = \theta,$ s.t. $\mathcal{P}^A_\theta \in \mathfrak{P}^A\}$;
	\item $\mathcal{A}^{\mu} = \bigcup_{k=1 \ldots N_\theta} X_k$;
	\item $\mathcal{P}^{\mu}$ is defined that for the current state $v_k = \langle x_1, x_2, \ldots , x_k \rangle$ and an action $x_{k+1}$, if $x_{k+1} \in X_{k+1}$ and $v_{k+1} = \langle x_1, x_2, \ldots , x_{k+1} \rangle \in \mathcal{S}^{\mu}$ the next state is $v_{k+1}$, otherwise $v_k$;
	\item $\mathcal{R}^{\mu}$ is defined that for terminal state $v_{N_\theta} = \langle x_1, x_2, \ldots , x_{N_\theta} \rangle = \theta$ the reward is the opposite number of the averaged return got by $\pi^A_{\phi^*}$ acting in $\mathcal{M}^A_\theta$, otherwise the reward is $0$.
\end{itemize}

In addition, the start state is $v_0 = \langle \rangle$ and the terminal states are $v_{N_\theta}= \langle x_1, x_2, \ldots , x_{N_\theta} \rangle$. Corresponding to this $\mathcal{M}^{\mu}$, $\pi^{\mu}_w(x_{k+1}|v_k;w)$ is designed to take an action $x_{k+1} \in X_{k+1}$ depending on the previous generated sequence $v_k$, and the transforming function $f^{\mu}$ is designed as $f^{\mu}(H^{\mu}) = v_{N_\theta} = \theta$. 
Note that due to the definition of $\mathcal{S}^{\mu}$, any partial parameter $v_t$ without potential to be completed as a valid parameter $\theta$ is avoided to be generated. This ensures any constraint on environment parameter can be followed. On the other hand, any valid $\theta$ is probable to be generated once $\pi^{\mu}_w$ is exploratory and of enough expression capacity.\footnote{The generative framework could also be applied for continuous environment generation although it results in low efficiency comparing to directly updating the environment by gradient.} 


\begin{figure}
	\centering
	\captionsetup[subfigure]{}
	\includegraphics[width=0.42\textwidth]{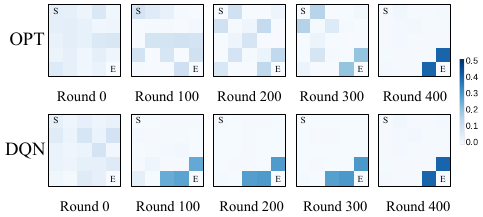}	
	\vspace{-5pt}
	\caption{Heatmaps of the blockage probability (soft wall, indicated by the intensity of the color in the cell) distribution throughout $5 \times 5$ soft wall Maze learning against the OPT and DQN agents.} \label{fig:soft-maze}
	\vspace{-10pt}
\end{figure}

\section{Experiments with Maze Design} \label{sec:maze}

\subsection{Experiment Setting}

\begin{figure} 
	\centering
	\includegraphics[width=0.38\textwidth]{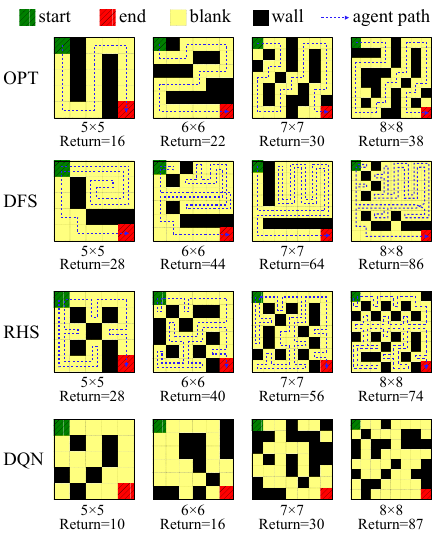}
	\vspace{-5pt}
	\caption{Best Mazes against OPT, DFS, RHS and DQN agents with size ranging from $5 \times 5$ to $8 \times 8$.}\vspace{-0pt}	\label{fig:best-result}
	\vspace{-10pt}
\end{figure}

\begin{figure} 
	\centering
	\includegraphics[width=0.42\textwidth]{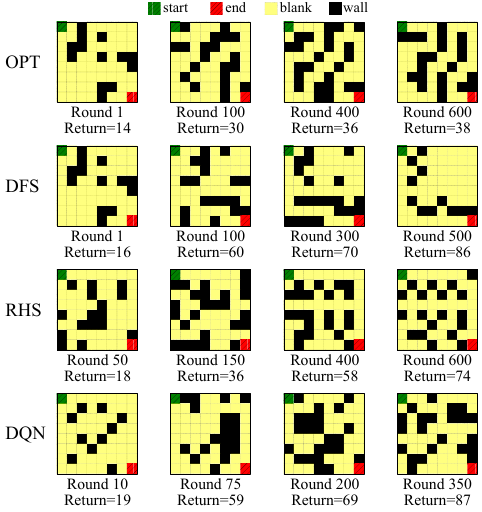}
	\vspace{-5pt}
	\caption{Learning to design Mazes against OPT, DFS, RHS and DQN agents in $8 \times 8$ map.} \label{fig:maze-evolving}
	\vspace{-10pt}
\end{figure}

In our experiment, we consider a use case of designing Maze game to test our solutions over the transition gradient method and the generative framework respectively. As shown in both Figs.~\ref{fig:best-result} and \ref{fig:maze-evolving}, the Maze is a grid world containing a map of $n \times n$ cells. 
In every time-step, the agent is in a cell and has four directional actions $\{ N, S, W, E \}$ to select from, and transitions are made deterministically to an adjacent cell, unless there is a \emph{wall} (e.g., the black cells as illustrated in Figs. \ref{fig:best-result} and \ref{fig:maze-evolving}), in which case no movement occurs. 
The minimax game is defined as: the agent should go from the north-west cell to the south-east cell using steps as few as possible, while the goal of the Maze environment is to arrange the walls in order to maximize the number of steps taken by the agent. 

Note that the above \emph{hard wall} Maze results in an environment that is discontinuous. In order to also test the case of continuous environments, we consider a \emph{soft wall} Maze as shown in Fig.~\ref{fig:soft-maze}. Specifically, instead of a hard wall that completely blocks the agent, each cell except the end cell has a blockage probability (soft wall) which determines how likely the agent will be blocked by this cell when it takes transition action from an adjacent cell. It is also ensured that the sum of blockage probabilities of all cells is $1$ and the maximum blockage probability for each cell is $0.5$. Thus, the task for the adversarial environment in this case is to allocate the soft wall to each cell to block the agent the most. 


Our experiment is conducted on PCs with common CPUs. We implement our experiment environment using Keras-RL \cite{plappert2016kerasrl} backed by Keras and Tensorflow. \footnote{Our experiment is repeatable and the code is at goo.gl/o9MrDN.}

\subsection{Results for the Transition Gradient Method}

We test the transition gradient method considering the $5 \times 5$ soft wall Maze case. 
We model the transition probability function by a deep convolutional neural network, which is updated by the transition gradient following Eq.~(\ref{eq:transition-gradient}). We consider the two types of agents: \textit{Optimal (OPT) agent} and \textit{Deep Q-network learning (DQN) agent}. The OPT agent has no parameters to learn, but always finds the optimal policy against any generated environment. The DQN agent \cite{mnih2013playing} is a learnable one, in which the agent's action-value function is modeled by a deep neural network, which takes the whole map and its current position as input, processed by 3 convolutional layers and 1 dense layer, then outputs the Q-values over the four directions. For each updated environment, we train the DQN agent to be optimal, as Fig.~\ref{fig:env-gan-iteration} shows.

Fig.~\ref{fig:soft-maze} shows the convergence that our transition gradient method has achieved. The change of the learned environment parameters, in the form of blockage probabilities, over time are indicated by the color intensity. Intuitively, the most effective adversarial environment to block the agent is to place two $0.5$ soft walls in the two cells next to the end or the beginning cell, as this would have the highest blockage probabilities. We can see that in both cases, using the OPT agent and the DQN agent, our learning method can obtain one of the two most optimal Maze environments. 

\subsection{Results for Generative Framework}

We now test our reinforcement learning generator by the hard wall Maze environment. We follow the proposed general generative framework to design $\mu_w = \langle \mathcal{M}^{\mu}, \pi^{\mu}_w, f^{\mu} \rangle$, which gradually generates walls one by one from an empty map. Particularly, $\pi^{\mu}_w$ is modeled by a deep neural network that takes an on-going generated map as input and outputs a position for a new wall or a special action for termination. Actions lead to generating walls that completely block the agent are invalid and prevented. We test our generator against four types of agents each on four sizes of maps (from $5 \times 5$ to $8 \times 8$). Although the objective for every agent is to minimize the number of steps, not every agent has the ability to find the optimal policy because of model restrictions of $\pi_\phi$ or limitations in the training phase. Therefore, besides testing our generator against the optimal agent (the OPT agent) and the DQN agent, we also adopt other two imperfect agents for our generator to design specific Mazes in order to understand more about our solution's behaviors. They are:


\textit{Depth-first search (DFS) agent.} The DFS agent searches the end in a depth-first way. In each time-step, without loss of generality, the DFS agent is set to select an action according to the priority of East, South, North, West. The DFS agent takes the highest priority action that leads to a blank and unvisited cell. If there are none, The DFS agent goes back to the cell from which it comes. 

\textit{Right-hand search (RHS) agent.} The RHS agent is aware of the heading direction and follows a strategy that always ensures its right-hand cell is a wall or the border. In each time-step, (i) the RHS agent checks its right-hand cell, if it is blank, the RHS agent will turn right and step into the cell; (ii) if not, then if the front cell is blank, the RHS agent will step forward; (iii) if the front cell is not blank, the RHS agent will continue turning left until it faces a blank cell, then steps into that cell.

Note that DFS and RHS are designed particularly for discontinuous Mazes. We also limit the network capacity and training time of the DQN agent to make it converge differently from the OPT agent. The learned optimal Mazes are given in Fig.~\ref{fig:best-result} for different agents with different Maze sizes. The strongest Mazes designed by our generator are found when playing against the OPT agent, shown in Fig.~\ref{fig:best-result} (OPT). We see that in all cases, from $5 \times 5$ to $8 \times 8$, our generator tends to design long narrow paths without any fork, which makes the optimal paths the longest. By contrast, the generator designs many forks to trap the DQN agent, shown in Fig.~\ref{fig:best-result} (DQN), as the DQN agent runs a stochastic policy ($\epsilon$-greedy).

In fact our generator could make use of the weakness from the agents to design the maps against them.
Fig.~\ref{fig:best-result} (DFS) shows the results that our generator designs extremely broad areas with only one entrance for the DFS agent to search exhaustively (visit every cell in the closed area twice). Fig.~\ref{fig:best-result} (RHS) shows the Mazes generated to trouble the RHS agent the most by creating a highly symmetric Maze.

Next, Fig.~\ref{fig:maze-evolving} shows the snapshots of the results in different learning rounds.
They all evolve differently, depending on the types of the agents. For the OPT agent, we find that our generator gradually links isolated walls to form a narrow but long path. For the DFS, our generator gradually encloses an area then broadens and sweeps it in order to best play against the policy that has the priority order of their travel directions. Fig.~\ref{fig:maze-evolving} (RHS) shows that our generator learns to adjust the wall into zigzag shapes to trouble the RHS agent. For the DQN agent, with limited network capacity or limited training time, it is usually the case that it cannot perfectly tell which road to go during the learning. As such, the generator tends to generate many forks to confuse the DQN agent.

\begin{figure}
	\centering
	\includegraphics[width=0.4\textwidth]{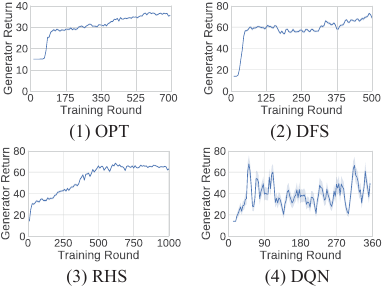}
	\vspace{-5pt}
	\caption{Training curves for OPT, DFS, RHS and DQN agents in $8 \times 8$ map. The lines and the shadows show mean and variance of generator return respectively.} \label{fig:training-curv}
	\vspace{-10pt}
\end{figure}

Furthermore, Fig.~\ref{fig:training-curv} shows the process of training our generator against the four agents in $8 \times 8$ map. We find that for OPT, DFS and RHS agents, the generator learns rapidly at first and gradually converges. But for the DQN agent, the learning curve is tortuous. This is because the ability of the DQN agent is gradually improved so it does not accurately and efficiently guide the learning of the generator. Also when the ability of the DQN agent improves greatly and suddenly, the learning curve for the generator may change its direction temporarily. Theoretically, training the DQN agent adequately in each iteration is a promising way towards to monotony and convergence. 


\section{Conclusions}

In this paper, we presented an extension of standard reinforcement learning by considering that the environment is strategic and can be learned. We derived a gradient method by introducing a dual MDP-policy pair for continuous environment. To deal with discontinuous environment, we proposed a novel generative framework using reinforcement learning. We evaluated the effectiveness of our solution by considering designing a Maze game. The experiments showed that our methods can make use of the weaknesses of agents to learn the environment effectively. 

In the future, we plan to apply the proposed methods to practical environment design tasks, such as video game design \cite{hom2007automatic}, shopping space design \cite{penn2005complexity} and bots routine planning. 

\section*{Acknowledgements}

This work is financially supported by National Natural Science Foundation of China (61632017) and National Key Research and Development Plan (2017YFB1001904).

\bibliographystyle{named}
\bibliography{ijcai18}

\end{document}